\documentclass[conference]{IEEEtran}
\IEEEoverridecommandlockouts

\usepackage{amsmath,amssymb,amsfonts}
\usepackage{algorithmic}
\usepackage{graphicx}
\usepackage{textcomp}
\usepackage{xcolor}
\usepackage{authblk}
\usepackage{tikz} 
\usetikzlibrary{shapes.geometric,arrows}
\usepackage{cite}
\usepackage{url}
\usepackage[hidelinks,colorlinks=true,linkcolor=blue,citecolor=green,urlcolor=blue]{hyperref}
\def\BibTeX{{\rm B\kern-.05em{\sc i\kern-.025em b}\kern-.08em
    T\kern-.1667em\lower.7ex\hbox{E}\kern-.125emX}}

\begin{document}
\title{Object Detection in Indian Food Platters using Transfer Learning with YOLOv4}

\author[1]{Deepanshu Pandey\textsuperscript{$\dagger$}}
\author[2]{Purva Parmar\textsuperscript{$\dagger$}}
\author[3]{Gauri Toshniwal\textsuperscript{$\dagger$}}
\author[4]{Mansi Goel\textsuperscript{$\dagger$}}
\author[5]{\\Vishesh Agrawal\textsuperscript{$\dagger$}}
\author[6]{Shivangi Dhiman\textsuperscript{$\dagger$}}
\author[7]{Lavanya Gupta}
\author[*]{Ganesh Bagler\textsuperscript{4}}

\affil[1]{\textit{\ ZS Associates India Private Limited, New Delhi, India}}
\affil[2]{\textit{\ Indian Institute of Science Education and Research (IISER) Pune, India}}
\affil[3]{\textit{\ Findability Sciences Inc., India}}
\affil[4]{\textit{\ Center for Computational Biology, Indraprastha Institute of Information Technology (IIIT Delhi), India}}
\affil[5]{\textit{\ Department of Computer Science, Indraprastha Institute of Information Technology (IIIT Delhi), India}}
\affil[6]{\textit{\ Department of Computer Science and Applied Mathematics, IIIT Delhi, India}}
\affil[7]{\textit{\ School of Computer Science, Carnegie Mellon University, Pittsburgh, PA, USA}}
\affil[*]{\textit{\ Corresponding author: Ganesh Bagler, bagler@iiitd.ac.in}

}
\renewcommand\Authands{ and }

\maketitle

\def\thefootnote{$\dagger$}\footnotetext{These authors contributed equally to this work}

\begin{abstract}
Object detection is a well-known problem in computer vision. Despite this, its usage and pervasiveness in the traditional Indian food dishes has been limited. Particularly, recognizing Indian food dishes present in a single photo is challenging due to three reasons: 1. Lack of annotated Indian food datasets 2. Non-distinct boundaries between the dishes 3. High intra-class variation. We solve these issues by providing a comprehensively labelled Indian food dataset- \emph{IndianFood10}, which contains 10 food classes that appear frequently in a staple Indian meal and using transfer learning with YOLOv4 object detector model. Our model is able to achieve an overall mAP score of 91.8\% and f1-score of 0.90 for our 10 class dataset. We also provide an extension of our 10 class dataset- \emph{IndianFood20}, which contains 10 more traditional Indian food classes. \\
\end{abstract}



\section{INTRODUCTION}
With the advent of deep learning, increasing computational power, and massive datasets, we have been able to achieve better accuracy in many domains. Even though deep learning is applied widely in the food industry, object detection on traditional Indian cuisine remains uncharted territory.

\begin{figure}[!hbt]
    \centering
    \includegraphics[width=2.5in]{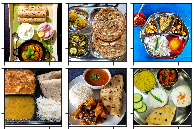}
    \caption{Some examples of Indian platter dishes}
    \label{Indian_Dish}
\end{figure}

A food photo, especially of an Indian platter (also called a \textit{thali}), comprises of several kinds of food dishes (see Fig.~\ref{Indian_Dish}). These food dishes can either be served in the same plate (non-distinct boundaries) or in different plates and bowls (distinct boundaries). In these cases, a single-label image classification models fail.

Multi-label object detection models such as F-RCNN, SSD and YOLO can detect multiple items in an image and localize them using bounding boxes. Localization is useful in developing real-world applications, wherein the user can find the position of each dish on a given plate. 

In this paper, we provide two labelled data sets: \textbf{\emph{IndianFood10}}: a dataset of 10 traditional Indian food items with more than 12,000 images, and \textbf{\emph{IndianFood20}}: an extension of IndianFood10 with 10 more popular Indian traditional Indian food classes. Further, we demonstrate the use of transfer learning with YOLOv4 on \emph{IndianFood10} dataset. Our key contributions include providing two comprehensive data sets and achieving a state-of-the-art mAP score of 91.8\% in object detection in the Indian food domain. Please note that our work with the 20 class data set is preliminary and we have not reported its results.

\section{RELATED WORK}
Several researchers addressed the challenges in the field of food image recognition. Kawano and Yanai~\cite{kawano2014food} used deep convolutional neural network to recognize food items from UECFood-100 image dataset~\cite{matsuda2012recognition} and achieved a top-1 accuracy of 72.26\% and top-5 accuracy of 92\%. In \cite{liu2016deepfood}, the authors used convolutional neural network on Food-101 and UEC-256 dataset, to achieve 77.4\% and 93.7\% as the top-1 and top-5 accuracy, respectively, on the former dataset. For the latter dataset, they reported 63.8\% and 87.2\% as the top-1 and top-5 accuracy, respectively. None of the above mentioned data sets had any Indian food item in them.\\

In the field of computer vision, object detection and localization are additional task aims to detect the food items from multiple food images. Previous studies used SIFT \cite{lowe1999object}, HOG \cite{wang2009evaluation} and SURF \cite{bay2006surf} on ImageNet and COCO datasets to detect the objects but could not achieve high accuracy. In recent years, deep learning models showed great improvement in object detection. Ashutosh et al. \cite{singla2016food} used deep neural network based GoogLeNet classifier to segregate food and non-food images and classified each food item. Amongst the initial works in food object detection, Matsuda et al.~ \cite{matsuda2012recognition} used various traditional computer vision methods for object detection on food images. In \cite{7900117} the authors used a very interesting technique for food item detection. They created a food activation map (probabilities heat map) and made bounding boxes using them and then used these boxes to classify the items in the boxes. Hoff et al.~\cite{hoff2018snap} have made a food object detection app which can be used to track user's daily food intake.\\

\begin{figure*}[!hbt]
    \centering
    \includegraphics[width=5.5in]{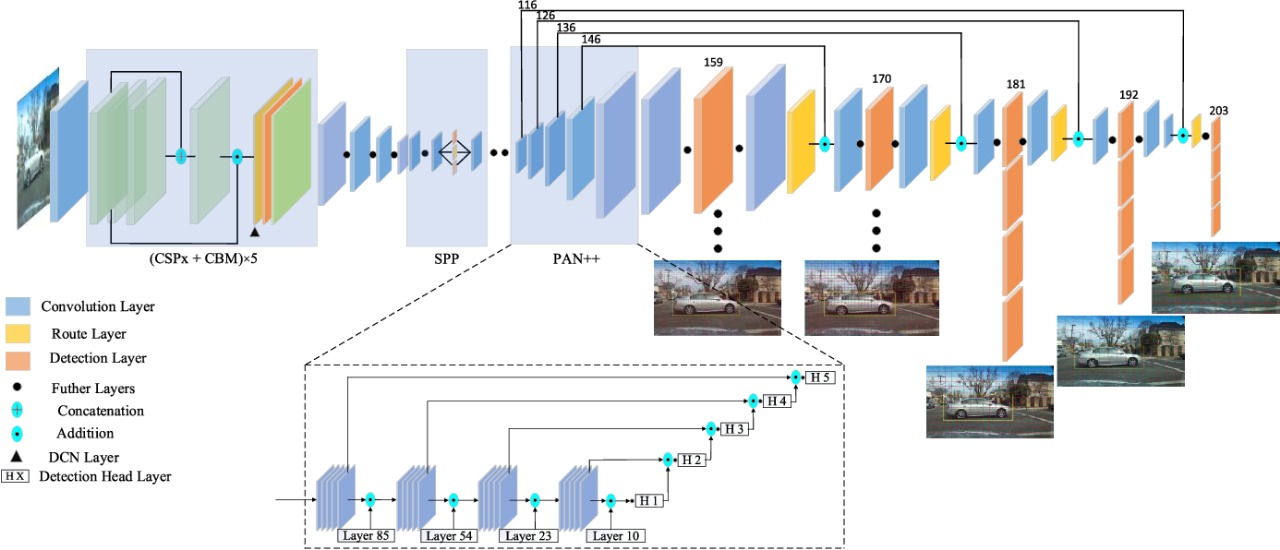}
    \caption{YOLOv4 Architecture \cite{yolo_diagram}}
    \label{yolo_arch}
\end{figure*}

Although some work is done in image recognition in Indian food context \cite{rajayogi2019indian,yadav2021food}, almost very little or no work is done in localising or detecting multiple Indian food items in an image. In \cite{ramesh2020real}, the authors performed object detection using Single Shot Detector(SSD) and Inceptionv2 on Indian food dataset for 60 classes using 4200 images (only 70 images per class). However, even after using image augmentation, their images per class is relatively low which will tend to overfitting and many of the food classes in their dataset were not traditional Indian dishes (e.g. pizza, pasta, noodles etc.) and some were very trivial classes (e.g. tomato, cucumber, water etc.). In BTBU-60 \cite{cai2019btbufood} the authors provide a dataset on 60 daily-use food items in Chinese cuisine for the object detection and some classes were mango, papaya, potato and tomato, and some Chinese items. Even though our objectives are similar (object detection in food items), their dataset domain is different than ours as our focus is on fully cooked Indian food cuisine and we do not have any raw ingredients or vegetables/fruits as part of our dataset.


As mentioned earlier, an Indian food platter (\emph{thali}, see Fig. 1) consists of several food items in a single plate and hence single image classification models are not useful for them. Therefore, two large datasets of 10 and 20 traditional Indian food items are proposed in this research along with the application of transfer learning with YOLOv4 for detecting multiple food items in a single image, which is suitable for Indian \emph{thali}.\\

\section{BACKGROUND}

\subsection{Object Detection}

Object Detection \cite{zou2019object} is a critical problem in the domain of computer vision where model's task is to locate and identify all objects in an image. Object Detection is a combination of object localization and classification for multiple objects in an image.\\ 

Earlier due to the lack of effective image representation techniques, handcrafted traditional methods namely Viola Jones Detectors, HOG Detector, Deformable Part-based Model (DPM) were used. With the advancement in technology, deep learning based methods namely Two Stage and One Stage Detectors are used to detect objects within an image.





\subsection{YOLOv4}

YOLO (You Only Look Once) is a fast one-stage object detector model. YOLO's architecture (Fig.~\ref{yolo_arch}) is similar to FCNN (Fully Convolutional Neural Network). Rather than just the local perspective, it considers the entire image and includes all the contextual information. It even uses features from the entire image to predict each bounding box.\\ 

\begin{figure*}[!hbt]
    \centering
    \includegraphics[width=18cm]{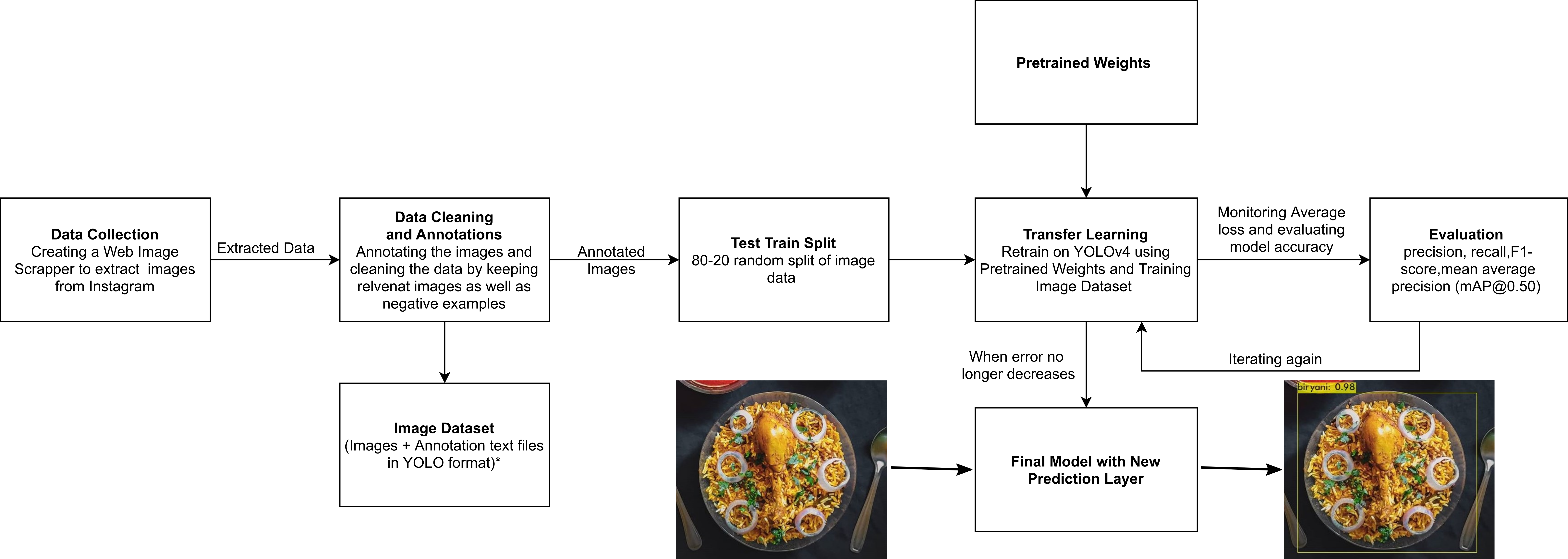}
    \caption{Flow chart of the steps and the model pipeline}
    \label{fig:pipeline}
\end{figure*}

YOLOv4 is the current state-of-the-art object detection model\cite{bochkovskiy2020yolov4} and was even used in detecting fashion apparel in \cite{lee2021two}. Starting with the backbone or the feature formation part, the authors go with CSPDarknet53 model based on DenseNet. It then contains an SPP block. Moving on, YOLOv4's Neck or the Feature Aggregation component is PANet. Finally the head or the detection step remains unchanged from the YOLOv3. It has 3 levels of detection along with an anchor based detection system.


\subsection{Metrics}

The metrics used for object detection is \textbf{Average Precision} (average of precision under different recalls) and \textbf{Mean Average Precision} (mAP: mean of average precision of different classes in the dataset). To find the object localization accuracy, \textbf{IoU} (Intersection Over Union) metric is calculated between the ground truth and the predicted bounding box. If the IoU is above the set threshold, then we infer that the object is successfully detected. Generally, the IoU threshold is set as 0.5. 

\begin{figure*}[!hbt]
    \centering
    \includegraphics[width=5.5in,height=7.5cm]{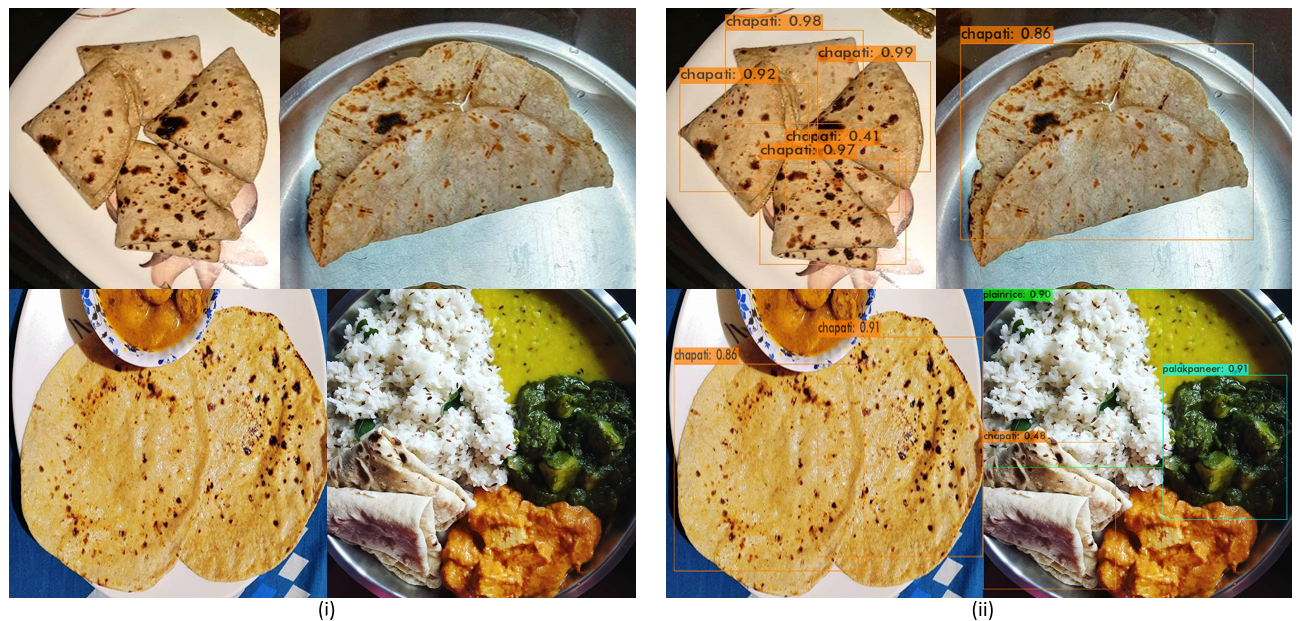}
    \caption{(i) Different orientations of chapati (ii) Model prediction}
    \label{fig:Chapati Image}
\end{figure*}

\section{MATERIALS AND METHODS}

\subsection{Data Preparation}
From a list of more than 100 Indian food items, we analyzed the number of images/posts on Instagram for each food item by using a hashtag and then selected few of the most popular Indian food items (such as \#alooparatha, \#plainrice, \#biryani etc.). We chose Instagram since it has more than 1 billion monthly users who share more than 100 million posts every day~\cite{nobles2020automated}. We used Python's \textit{Selenium} library to scrape Instagram post's URLs for every hashtag and then downloaded the images using another python library, \textit{Requests}, similar work was done by \cite{nobles2020automated} where they used \#HIV to download images from Instagram.

\subsection{Dataset}
Our \textit{IndianFood10} dataset consists of a total of 11,547 images with annotation text files for every image (see Table~\ref{tab:perf_stat} for all the food classes in \textit{IndianFood10}). Out of the total images, 842 images (approximately 7\% of the whole dataset) are multi-dish images (i.e., image containing more than one unique food class). For single dish images, we annotated the single dish of interest. For platter (multi-dish) images, we annotated more than one dish of interest. The average dishes-per-image ratio for platter images is 2.33.

By the virtue of being such a diverse cuisine, each Indian food dish can be paired with multiple other food dishes, yielding no clear boundaries between two dishes. Even the same dish may have large visual variance (high intra-class variation) because of different cooking methods and presentation, which brings certain challenges to the recognition \cite{Wang2019}. For example, the class \textit{chapati} (a type of Indian bread) can co-occur with various classes like \textit{palak paneer}, \textit{plain rice} etc. in different orientations (full-opened, half-folded and quarter-folded, (Fig.~\ref{fig:Chapati Image} (i)). Hence, a large number of images of each class at different scales, lighting, rotations, sides and on different backgrounds are required \cite{bochkovskiy2020yolov4} and due to this reason we have such a large dataset.

\subsection{Annotation}

We annotated the images using an open-source software, makesense.ai~\cite{make-sense}. Here, we uploaded the raw input images of our dataset and annotated each food item of interest present in every image by manually creating the bounding boxes and labelling each box with the food class. For every image in the dataset, a text file was generated in YOLOv4 format which contained the information about the coordinates of the bounding boxes created for each food item in an image with the food class number. 

\subsection{Approach}

We trained Alexey Bochkovskiy's YOLOv4\cite{bochkovskiy2020yolov4} object detection model on 80\% of the entire \textit{IndianFood10} dataset (single-dish images and platter images), the dataset is available on \href{https://drive.google.com/drive/u/0/folders/16kRxAgfQfVBD2ebLdHTqylZdWSB2rygQ}{Google Drive} and also uploaded on IEEE dataport \cite{IndianFoodDataset}. The model was trained on Google Colab, which provided Tesla K80 and Tesla T4 GPUs. At the end of training, the metrics were computed by testing against the validation set (20\% of the \textit{IndianFood10} dataset).

\section{RESULTS AND DISCUSSION}
The most common evaluation metrics for any object detection model are Precision-Recall curves and Average Precision. We use the standardised code provided by Padilla et al.~\cite{padillaCITE2020} to compute our model scores. With an IoU threshold of 0.5, we achieve a 10-class mean average precision (mAP) score of 91.76\% and F1 score of 0.90 (see Table~\ref{tab:perf_it}).

\begin{figure}[hbt!]
    \centering
    \includegraphics[width=3.25in]{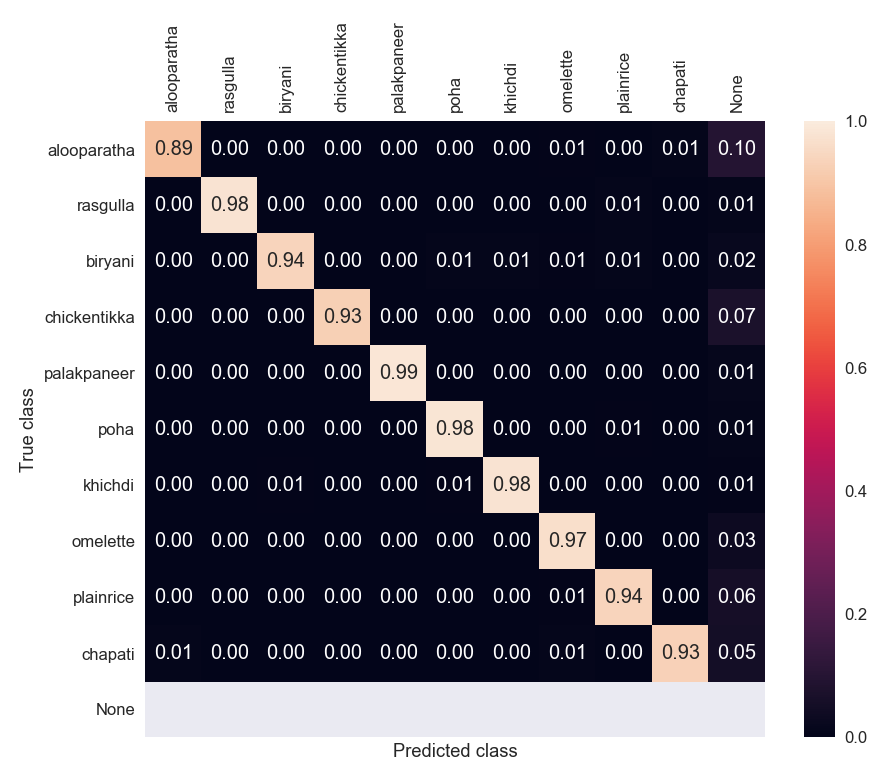}
    \caption{Confusion Matrix for 10 classes}
    \label{confusion_mat}
\end{figure}

\begin{figure*}[!hbt]
    \centering
    \includegraphics[width=15cm]{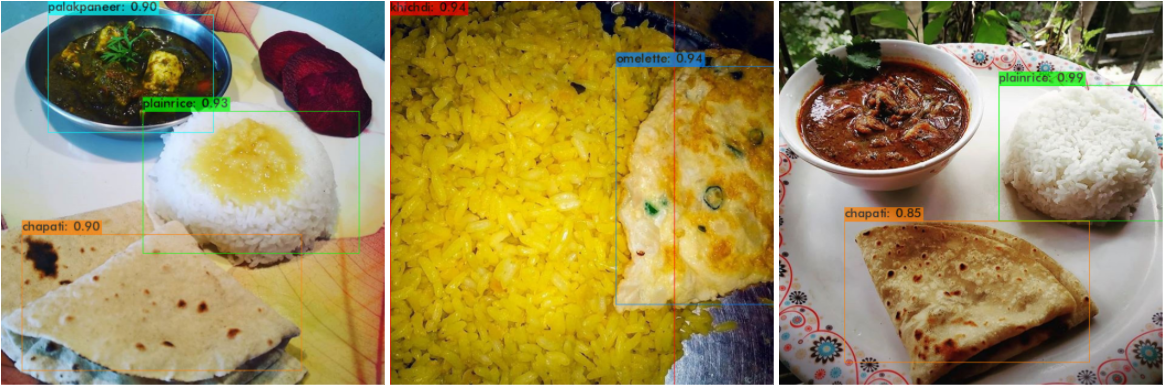}
    \caption{Predictions from trained YOLOv4 model}
    \label{fig:fig_results}
\end{figure*}

\begin{figure*}[!hbt]
    \centering
    \includegraphics[width=14cm]{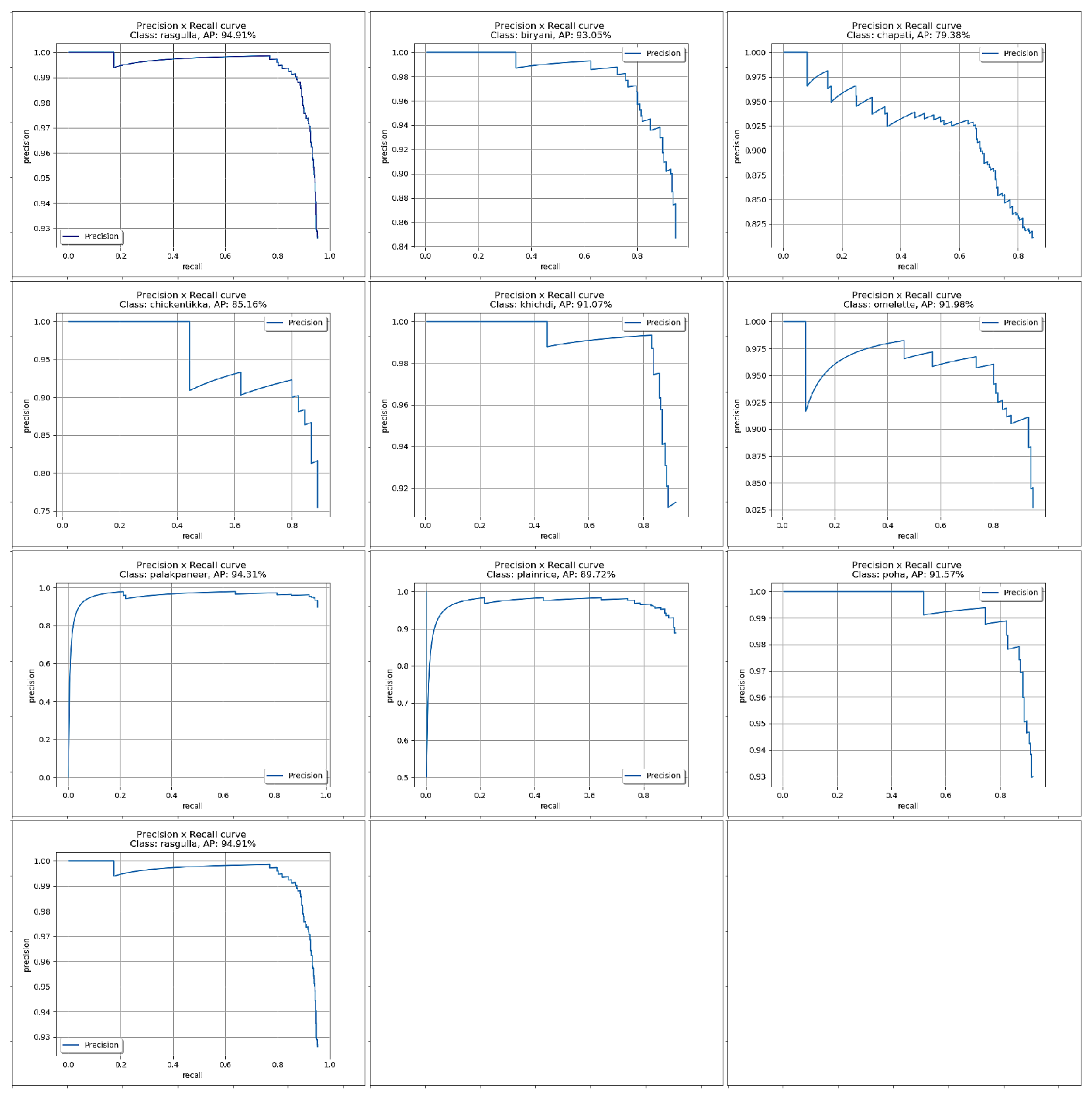}
    \caption{PR Curves for 10 classes}
    \label{PR_curve}
\end{figure*}

Table~\ref{tab:perf_stat} lists the individual class average precision scores and their confusion matrix (Fig.~\ref{confusion_mat}). An extra class \textit{None} was introduced to account for images where the model failed to predict any class. The true class of a single-dish image can never be \textit{None}, and hence the last row in the confusion matrix has been greyed out. Our model performs very well and it is able to detect food classes correctly despite many of the food items having high intra-class variation (see Fig.~\ref{fig:Chapati Image} (ii)) and no clear boundaries between them (see Fig.~\ref{fig:fig_results}). Fig.~\ref{PR_curve} shows the PR-curves for all the 10 classes.

\begin{table}[!hbt]
    \centering
    \caption{Average Precision for Each Class}
    \label{tab:perf_stat}
    \begin{tabular}{|l|c|} \hline
        \textbf{Class in \textit{IndianFood10}} & \textbf{Average Precision (AP) in \%} \\ \hline
        Aloo Paratha & 78.3 \\ 
        Biryani & 93.0 \\
        Chapati & 79.4 \\
        Chicken Tikka & 85.1 \\
        Khichdi & 91.0 \\
        Omelette & 91.9 \\
        Palak Paneer & 94.3 \\
        Plain rice & 89.7 \\
        Poha & 91.5 \\
        Rasgulla & 94.9 \\ \hline
    \end{tabular}
\end{table}

\begin{table}[!hbt]
    \centering
    \caption{Mean Average Precision for each iterations}
    \label{tab:perf_it}
    \begin{tabular}{|c|c|c|} \hline
        \textbf{Iterations} & \textbf{Mean Average Precision (in \%)} & \textbf{F1-Score} \\ \hline
        7000 & 90.49 & 0.89\\ 
        8000 & 91.57 & 0.90\\
        9000 & 90.75 & 0.89\\
        \textbf{10000} & \textbf{91.76} & \textbf{0.90}\\
        11000 & 90.99 & 0.90\\
        12000 & 90.80 & 0.90\\
        13000 & 91.03 & 0.90\\
        14000 & 90.41 & 0.90\\
        15000 & 90.26 & 0.90\\
        16000 & 90.28 & 0.90\\ 
        17000 & 90.83 & 0.91\\ 
        18000 & 89.89 & 0.90\\ 
        19000 & 90.16 & 0.91\\ 
        20000 & 90.83 & 0.91\\ 
        \hline
    \end{tabular}
\end{table}

\begin{table}[!hbt]
    \centering
    \caption{Summary of mAP Scores}
    \label{tab:comp_Score}
    \begin{tabular}{|l|c|} \hline
    \textbf{Model} & \textbf{mAP Score} \\ \hline
    BTBU-Food-60 \cite{cai2019btbufood} & 67.7\% \\
    SSD\_InceptionV2 \cite{ramesh2020real} & 76.9\% \\ 
    \textbf{YOLOv4 on \textit{IndianFood10}} & \textbf{91.8\%} \\ \hline
    \end{tabular}
\end{table}

\section{CONCLUSION}
Our literature survey overview revealed that there is a lack of work done on object detection in the context of Indian cuisine. We have been able to curate a large dataset (\textit{IndianFood10}) with more than 11,000 annotated images for 10 popular Indian dishes as classes. We achieved a mAP score of 91.8\% for object detection in Indian cuisine using the YOLOv4 architecture, Table~\ref{tab:comp_Score} summarises the mAP scores of previous research works done in the field of object detection in food items. This work has implications for calorie estimation in the food images and thus is expected to have a larger impact of public health.


\section{FUTURE WORK}
With the rise of computer vision, object detection continues to be an important problem and especially in traditional Indian food context because currently there is no public dataset available. As to scale up our proposed work and dataset, we have also created a dataset with 20 Indian dishes as classes- \textit{IndianFood20} (an extension of \textit{IndianFood10}) which contains 17,817 images (see Table~\ref{tab:food20}). Our final work for \textit{IndianFood20} is preliminary but we would like to share the datasets \textit{IndianFood10} and \textit{IndianFood20} with the research community so that work on the area of object detection in context of Indian cuisine could be accelerated. Future research opportunities in Indian food context include:

\begin{itemize}
  \item Deploying a mobile application for detecting food items and provide its recipe, ingredients and nutrition facts \cite{Sun2019FoodTrackerAR}
  \item Estimation of total calories present in a meal by considering the volume of each food item present in it
\end{itemize}

\begin{table}[!hbt]
    \centering
    \caption{Food classes in \textit{IndianFood20}}
    \label{tab:food20}
    \begin{tabular}{|l|l|} \hline
    \multicolumn{2}{|c|}{\textbf{List of Food Items}}  \\ \hline
        Indian Bread & Dosa \\ 
        Rasgulla & Rajma \\
        Biryani & Poori \\
        Uttapam & Chole \\
        Paneer & Dal \\
        Poha & Sambhar \\
        Khichdi & Papad \\
        Omelette & Gulab Jamun \\
        Plain Rice & Idli \\
        Dal Makhni & Vada \\ \hline
    \end{tabular}
\end{table}

\section*{ACKNOWLEDGEMENT}
G.B. thanks Indraprastha Institute of Information Technology (IIIT Delhi) for the computational support. G.B. thanks Technology Innovation Hub (TiH) Anubhuti for the research grant. D.P, P.P, G.T, V.A, S.D are summer interns and M.G. is a research scholar in Dr. Bagler's lab at IIIT Delhi and thankful to IIIT Delhi for the support. M.G. thanks IIIT Delhi for the fellowship.

{\small
\bibliographystyle{IEEEtran}
\bibliography{food_recog}
}

\end{document}